\pdfoutput=1

\documentclass[a4paper, 10pt, conference]{ieeeconf}      % Use this line for a4 paper

\IEEEoverridecommandlockouts                              % This command is only needed if 
\usepackage{graphicx} % for pdf, bitmapped graphics files
\usepackage{adjustbox,lipsum}
\usepackage{xcolor}
\usepackage{subfigure}
\usepackage{amsmath}
\usepackage{hyperref}\hypersetup{
    colorlinks=true,
    linkcolor=blue,
    filecolor=magenta,      
    urlcolor=cyan,
}

\title{\LARGE \bf Fast object detection in compressed JPEG Images}

\author{Benjamin Deguerre$^{1,2}$, Cl\'ement Chatelain$^{1}$, Gilles Gasso$^{1}$% <-this % stops a space
\thanks{$^{1}$ LITIS EA 4108, INSA Rouen, Normandie Universit\'e
        {\tt\small firstname.lastname@insa-rouen.fr}}%
\thanks{$^{2}$ ACTEMIUM Paris Transport
        }%
}
\begin{document}

\maketitle
\thispagestyle{empty}
\pagestyle{empty}

%%%%%%%%%%%%%%%%%%%%%%%%%%%%%%%%%%%%%%%%%%%%%%%%%%%%%%%%%%%%%%%%%%%%%%%%%%%%%%%%
\begin{abstract}

Object detection in still images has drawn a lot of attention over past few years, and with the advent of Deep Learning impressive performances have been achieved with numerous industrial applications. Most of these deep learning models rely on RGB images to localize and identify objects in the image. However in some application \emph{scenarii}, images are compressed either for storage savings or fast transmission. Therefore a time consuming image decompression step is compulsory in order to apply the aforementioned deep models. To alleviate this drawback, we propose a fast deep architecture for object detection in JPEG images, one of the most widespread compression format.  We train a neural network to detect objects based on the blockwise DCT (discrete cosine transform) coefficients {issued from} the JPEG compression algorithm. We modify the well-known Single Shot multibox Detector (SSD) by replacing its  first layers with one convolutional layer dedicated to process the DCT inputs. Experimental evaluations on PASCAL VOC and industrial dataset comprising images of road traffic surveillance show that the model is about $2\times$ faster than regular SSD with promising detection performances.  To the best of our knowledge, this paper is the first to address detection in compressed JPEG images.
\end{abstract}

 %at  the  expense  of  a  reasonable  decrease  in detection performa

%%%%%%%%%%%%%%%%%%%%%%%%%%%%%%%%%%%%%%%%%%%%%%%%%%%%%%%%%%%%%%%%%%%%%%%%%%%%%%%%

\section{INTRODUCTION}

Deep convolutional networks are now the reference models in image classification or object detection tasks \cite{gu2018recent,agarwal2018recent}. The achieved levels of performance and robustness allow for usage of deep models in numerous application domains including medical imaging,  classification and detection on aerial imagery,  autonomous driving, road surveillance systems. In many contexts either for storage or fast transmission purposes,  it is common to compress the images - or videos they are extracted from - in formats such as JPEG, JPEG 2000, PNG, GIF for images, and mpeg4 part2 or H.264 for videos.
Some of the most popular compression algorithms such as JPEG, mpeg4 part2 or H.264 share a common encoding strategy based on a block transformation of the images before entropy coding. Applying generic convolutional neural networks (CNN) would require a decoding step to RGB format, which is computationally costly and memory demanding.

Thereon, a recent trend of research work aims at processing directly compressed images  or videos in order to speed up training and inference of deep networks. For image classification, Ulicny et al. (2007) \cite{ulicny2017using} learn CNNs either with raw YCbCr image representation or its compressed DCT (Discrete Cosine Transform) blocks. A step was taken further by Gueguen et al. (2018) \cite{uber_classification_dct} who modify a ResNet-50 network \cite{resnet} to account for the different resolutions of Y and Cb/Cr DCT blocks. The resulting networks are 1.77$\times$ faster at inference and attain state-of-the-art classification performances. Another research stream designs dedicated networks to spectral input coefficients: harmonic networks \cite{ulicny2018harmonic} uses custom convolutions that produce high-level features by learning combinations of spectral filters defined by the 2D Discrete Cosine Transform; Ehrlich and Davis (2019) \cite{ehrlich2018deep} introduce a ResNet able to operate on compressed JPEG images by including the compression transform into the network weights. From video side, two recent works on detection in compressed videos are \cite{wu2018compressed, wang2018fast}. In \cite{wu2018compressed}, separate CNNs are used for temporally linked I-frame (RGB image), and P-frame (motion  and residual arrays) are trained all together. In \cite{wang2018fast}, the authors consider three networks: a CNN feature extraction module based on the raw I-image, a recurrent memory network to align the features of consecutive P-frames using compressed motion and residual vectors, and a detection network aiming at identifying the objects in the videos.

Inspiring from this body of research on compressed images, this paper addresses object detection problem in compressed JPEG images with application to road surveillance images. Contrary to classification task, detection in that setting raises the question of spatial localization of the objects in the frequency domain. We show that detection in frequency domain is feasible and lead to substantial speed improvement with promising detection performances. The gain is twofold: i) the full decoding steps of JPEG image is no longer required as our approach only requires extraction of DCT coefficient blocks, and ii) due to the block-wise nature of the inputs, the first convolution layer reduces by a factor of 8 the inputs size $(w , h) \rightarrow (\frac{w}{8}, \frac{h}{8})$, speeding up the whole network by a factor 2. We also provide insights about the detection capacity of the network and show that the proposed approach may struggle in detecting small size objects.

The remainder of the article is organized as follows: section 2 describes related work and emphasizes on object detection in RGB images. It also introduces the JPEG encoding norm and the way it can be leveraged on to design or fast detector. Section 3 presents the proposed approach. Experimental evaluations and drawn remarks are deferred to Section 4.

\section{RELATED WORKS}

\subsection{Object detection}

Object detection using deep neural networks has been widely studied, and recently many efficient detectors were proposed. 
%such as \cite{rcnn}, \cite{fast_rcnn}, \cite{faster_rcnn}, \cite{mask_rcnn}, \cite{rfcn}, \cite{ssd}, \cite{yolo}, \cite{yolov2}, \cite{yolov3}, \cite{fssd} exists. 
R-CNN \cite{rcnn} and successive improved versions, Fast R-CNN \cite{fast_rcnn} and Faster R-CNN \cite{faster_rcnn} use a Region Proposal Network (RPN) based on candidate object propositions in the image, followed by a classification step to infer the class of each box. 
%Each of the architecture is an improvement of the previous one (starting from \cite{rcnn}). 
Mask R-CNN \cite{mask_rcnn} extends the framework of Faster R-CNN by adopting a similar region of interest (ROI) proposal followed by object class prediction, bounding box regression and an additional task of producing binary mask for each ROI. This leads to a performance boost both in segmentation and detection. The R-FCN \cite{rfcn} uses a similar pattern replacing the final classification by pooling and vote to keep the network fully convolutional, thus improving the speed of the network. 

Another prevalent family of detectors relies on one-stage approach to get faster detection rate. YOLO \cite{yolo}, and subsequent upgrades YOLOv2 \cite{yolov2} and YOLOv3 \cite{yolov3} divide an image into a grid of $S \times S$, and every grid predicts $N$ bounding boxes with confidence scores. The score accesses the precision of predicted boxes and object class.  As the prediction is based on image global features issued from convolutional layers, YOLO \cite{yolo} greatly improves the detection speed, at the cost of detection precision. Nonetheless, YOLOv3 can almost reach performances of two-stage methods by integrating several improvements, such as multilabel object class prediction, prediction across scales or the use of K-means clustering to determine box priors.  Another popular one-shot detector is SSD \cite{ssd}. SSD considers a fixed set of default bounding boxes at different scales and aspect ratios with associated feature map. By coupling box matching strategy with the multi-scale features, SSD is significantly more accurate than YOLO \cite{yolo} with interesting detection speed.  FSSD \cite{fssd} improves on SSD by  adding a lightweight feature fusion module to combine the multi-scale feature maps of SDD and then by generating feature pyramid to predict the boxes. This allows a better detection of small objects.

\subsection{Classification backbones of object detectors}
Most of the deep neural networks for detection rely on existing classification modules except YOLO and its variants \cite{yolo,yolov2,yolov3} which directly pre-train their networks on Imagenet. For instance, detectors such as Fast R-CNN \cite{rcnn}, Faster R-CNN \cite{faster_rcnn}, SSD \cite{ssd} or  FSSD \cite{fssd} use the VGG16 network \cite{vgg} a as backbone, while R-FCN \cite{rfcn} is based on deep residual network \cite{resnet}.  VGG16 operates a series of convolution layers followed by 3 densely connected layers, while Residual network introduces a residual learning strategy to implement deeper network. Note that although deeper network than VGG16, ResNet is lighter in term of FLOPs. 

The object detection networks are not linked to a specific classification module, and can easily use other classification backbones, provided the dimensions of input images and network outputs are adjusted accordingly. 

\subsection{Neural networks and Compression}
The detection networks we discussed share a common feature: they all act on plain RGB images. To the best of our knowledge, few research works investigate a direct use of compressed images either for classification or detection. The most relevant work in the domain is \cite{uber_classification_dct}, where JPEG images are partially decoded to extract $8 \times 8$  DCT coefficients blocks. These frequency domain features are then fed to a modified Resnet50 network for classification task. Because of the $8 \times 8$  DCT blocks, the original shape of the input is divided by 8: $(w , h, 3) \rightarrow (\frac{w}{8}, \frac{h}{8}, 3 \times 8 \times 8)$. It allows to skip the first computationally expensive convolution layers of the original RGB-based architecture, leading to 1.77$\times$ speed up at inference stage with classification performance slightly superior to RGB-based networks. 

For object detection task we are aware of the approach of Torfasson et al. (2018) \cite{torfason2018towards} who design an encoder-decoder neural network to learn compressed representations of RGB images. These representations are further used to trained deep convolutional networks for classification or semantic segmentation. This approach \cite{torfason2018towards} is most related to the one we propose hereafter except the fact  we exploit the readily acquired JPEG compressed images.

\subsection{JPEG Norm}
\begin{figure}[!ht]
    \centering\includegraphics[width=\columnwidth]{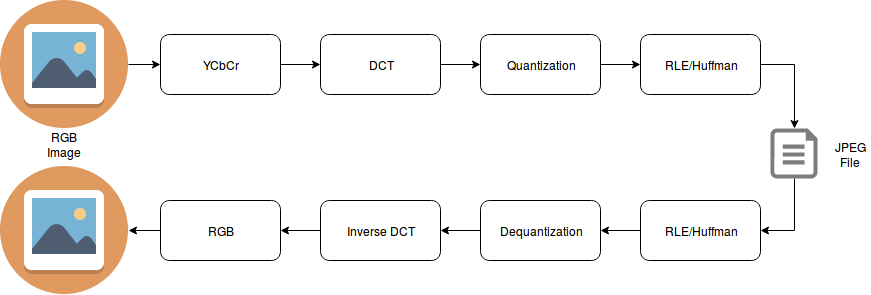}
    \caption{JPEG CODEC pipeline}
    \label{fig:JPEG_pipeline}
\end{figure}

JPEG encoding and decoding pipeline is summarized in  Figure \ref{fig:JPEG_pipeline}.  When going through compression, the following steps are applied:

\begin{itemize}
    \item The image is first converted to YCbCr and subsampled
    \item Then a block-wise DCT is applied
    \item A block-wise quantization is applied
    \item An RLE/Huffman compression algorithms are applied for the entropy coding
\end{itemize}

\begin{figure}[!ht]
    \centering
    \includegraphics[scale=0.75]{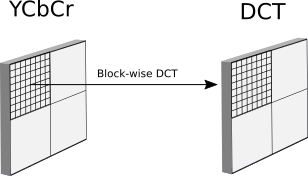}
    \caption{Block-wise DCT example. The Block of YCbCr coefficients is converted to the frequency domain using a 2D Discrete Cosine Transform.}
    \label{fig:blockwise_dct}
\end{figure}

Up until the RLE/Huffman coding, the data is image based, it keeps the same shape as the input image (modulo the sub-sampling), after the RLE/Huffman encoding the shape of the data will vary from image to image. Depending on the compression ratio specified to the encoding algorithm, the subsampling ratio may change from one image to an other. The quantization coefficients may also vary between different encoding processes. The transformation from RGB to YCbCr and the block-wise DCT are on the other hand fixed by equations for all the images. Figure \ref{fig:blockwise_dct} gives a visual understanding of the block-wise DCT as this is the key feature that allows for detection speed improvements and the function to calculate the DCT is given in the equation \ref{eq:fdtc}.

\begin{equation}\label{eq:fdtc}
        S_{uv} = \frac{1}{4} C_u C_v \sum_{x=0}^7 \sum_{y=0}^7 s_{yx} cos \frac{(2x+1)u \pi}{16} cos \frac{(2y+1)v \pi}{16} \\
\end{equation}
where:
\begin{equation*}
    C_u, C_v =
    \begin{cases}
       \frac{1}{2}, & \text{for }\  u, v = 0 \\
       1, & \text{otherwise}
    \end{cases}
\end{equation*}

\section{Proposed approach}
\begin{figure}[!ht]
    \centering
    \includegraphics[width=0.9\columnwidth]{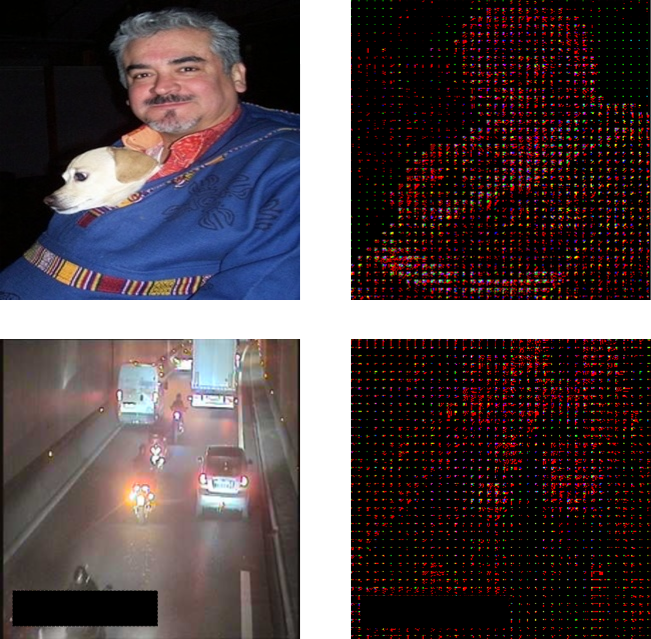}
    \caption{Illustration of block-wise Discrete Cosine transformation on two images. \textbf{Left}: RGB image; \textbf{Right}: array of $8\times 8$ blocks DCT coefficients. \textbf{First row}: Pascal VOC image; \textbf{Second row}: image from driving road (private dataset).}
    \label{fig:block_transform}
\end{figure}

The problem we address is the detection of objects from JPEG encoded images. We aim at identifying the class and spatial localization of objects of interest (car, truck, motorcycle, see second row of Figure \ref{fig:block_transform}) from a deep CNN, acting on the compressed frequency domain representation. To highlight the difficulty of the problem, Figure \ref{fig:block_transform} shows plain RGB images and corresponding compressed representations after Discrete Cosine Transformation.  We describe in the next subsections the designed network, its input and the learning problem it involves. 

\subsection{Design of the Network}
Many inputs can be considered for our detection network. Inspiring from \cite{uber_classification_dct} we partially decode the JPEG files up to the DCT coefficients in the middle of dequantization and inverse DCT steps (see Figure \ref{fig:JPEG_pipeline}). The main benefit is that for an RGB image $I$ of dimension $(h, w, 3)$, we get a fixed size input array $D$ downsized by the first convolution to a size of $(h/8, w/8, 3 \times 8 \times 8)$. In order to proof the concept, we do not consider situations where up-sampling and down-sampling operations are applied to YCbCr when encoding $I$. Instead we assume all images are re-sampled to a factor $4:4:4$ (no subsampling). Note that this can be easily generalized to any subsampling policies.

\begin{figure*}[!ht]
    \centering
    \begin{subfigure}{\includegraphics[width=0.75\linewidth]{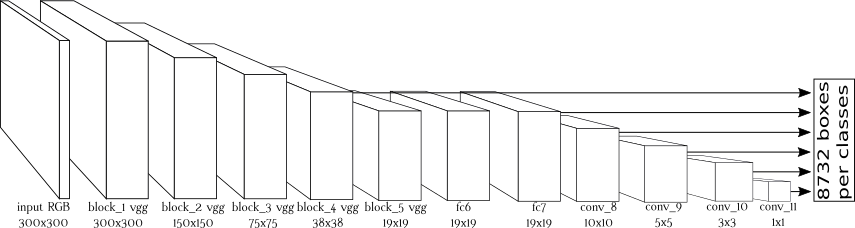}}
       \caption{Original SSD network in the RGB space.}
       \label{fig:Ng1} 
    \end{subfigure}

    \begin{subfigure}{\includegraphics[width=0.75\linewidth]{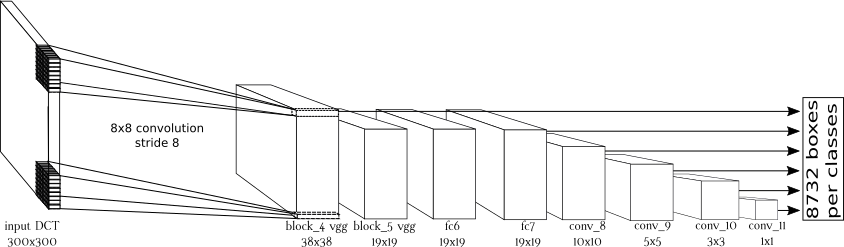}}
       \caption{Modified SSD network using partially decoded JPEG images as input. The four first blocks of the regular SSD is replaced by a lightweight convolution layer dedicated to process DCT blocks, using a $8\times8$ kernel and a stride of 8 to keep the consistency of the DCT blocks.}
       \label{fig:Ng2}
    \end{subfigure}
\end{figure*}

The proposed architecture is depicted on figure \ref{fig:Ng2}. We modified the original SSD architecture (Figure \ref{fig:Ng1}) in order to account for the blockwise nature of the input. As shown in Figure \ref{fig:blockwise_dct}, DCT is applied on $8 \times 8$ block of pixels at encoding stage. Therefore DCT input $X$ includes spatial information of original image $I$ at a lower resolution of factor $8 \times 8$. To preserve this spatial information, adjacent compressed blocks should not be mixed up, and therefore we have turned toward the use of a DCT-dedicated first convolution layer set with a filter size of $(8 \times 8)$ and a stride of 8. Figure \ref{fig:Ng2} depicts how each  DCT block is mapped into a single location within the VGG16 block4. The features at this location will be used to predict the anchor boxes at this position in the image, making the SSD an excellent candidate for the block-wise frequency input. Because we remove the first blocks from the VGG16 and sparsely apply the convolution to the input, the network speed greatly improves by removing a large number of dense convolutions on the biggest feature maps.

While we use the SSD for our experiments, many RGB detection networks could theoretically be used for detection with block-wise DCT input, provided it has the shape of the spatial image. The usage of other networks is out of the scope of this article and up for study.

\subsection{Training procedure}

As stated before, detection networks are built upon a classification neural network. Hence, training our proposed SSD architecture requires a classification model able to handle DCT inputs. Such off-the-shelves model being unavailable, we start by learning a dedicated backbone VGG model. Its architecture is as in Figure \ref{fig:Ng2}. The first three blocks of the network are removed and replaced by the convolution described above to fit the new input. We also normalize the input data.

Following the procedure described in \cite{vgg}, we first train a shallow classifier, namely VGG-A, on the ILSVRC 2012 dataset \cite{imagenet_dataset}. We then load the learned weights into a deeper version of the network, namely VGG-D (the usual VGG16) and re-train the network. For this training we re-use the same hyper-parameters as the ones used in \cite{vgg} but we do not carry the extensive data-augmentation described. We simply resize the images to $224 \times 224$ and input the images to the classification network. In order to fit the network architecture, we re-sampled before-hand all the images to 4:4:4 sampling ratio.

The results are described in Table \ref{table:vgg16_training}. As the ImageNet evaluation servers are not up anymore, the results described are on the validation dataset. All the results are rather far from the state of the art, but this is of little importance as we only want the weights to initialize our detection network. Nonetheless, we are wary of the fact that the low convergence might impede the training of the detection network, so we trained the VGG both in spatial and frequency domain to be able to do comparison.

\begin{table}[!ht]
    \centering
    \begin{adjustbox}{max width=\linewidth}
        \begin{tabular}{|l|c|c|c|}
        \hline
            Network &  Training Space & top-1 accuracy & top-5 accuracy \\
        \hline
            VGG-D \cite{vgg} & RGB & 75.6 (test) & 92.8 (test) \\
        \hline
            VGG-D & RGB & 49.8 (val) & 74.8 (val) \\
        \hline
            VGG-D (FS) & DCT & 40.3 (val) & 65.1 (val) \\
            VGG-D (FA) & DCT & 42.0 (val) & 66.9 (val) \\
        \hline
        \end{tabular}
    \end{adjustbox}
    \caption{Classification results for the training of the VGG (FS = from scratch, FA = from VGGA)}
    \label{table:vgg16_training}
\end{table}

These networks were then used as backbone for the SSD detection network. We re-use the SSD data augmentation techniques and all the hyper-parameters were set to match the one described in \cite{ssd}. We train the SSD using our trained weights for fair comparison between spatial and frequency domain inputs, we also give the results from the original SSD article. All the results are described in the next section, in the table \ref{table:ssd_results_pascal}.

\section{Experiments and Results}

\subsection{Datasets and evaluation metrics}

We carried out experiments on two datasets, the public PASCAL VOC dataset \cite{pascalvoc} and a private one, referred to as ACTEMIUM dataset. PASCAL VOC data are composed of 11,530 natural scene images containing a total of 20 classes with bounding boxes for each object. We create 2 training sets by combining the data available: `07` for the PASCAL VOC 2007 train-validation dataset and `07+12` for the union of the PASCAL VOC 2007 train-validation and 2012 train-validation dataset. ACTEMIUM dataset includes images taken from different cameras of a video surveillance system intended to monitor road traffic in tunnels in Paris (France) area. The dataset contains 3 classes (car, truck, motorcycle) with their bounding boxes and is randomly split into 1578 training images, 380 validation images and 218 test ones.  {The class distribution is detailed in table \ref{table:miisst_class_detail}}.

\begin{table}[!ht]
    \centering
    \begin{adjustbox}{max width=\linewidth}
        \begin{tabular}{|c|c|c|c|c|c|c|c|}
        \hline
            Set & Number of images & car & truck & motorcycle \\
        \hline
            training & 1578 & 4303 & 658 & 142 \\
        \hline
            validation & 380 & 1012 & 143 & 22 \\
        \hline
            test & 218 & 588 & 79 & 29 \\
        \hline
        \end{tabular}
    \end{adjustbox}
    \caption{Class distribution per set in ACTEMIUM dataset}
    \label{table:miisst_class_detail}
\end{table}

For each dataset, we trained at least 2 SSD-based detection models:
\begin{itemize}
    \item a detector for compressed images, referred to as SSD\_freq, and
    \item a second RGB input SSD initialized with our pre-trained VGG model, hereafter denoted SSD\_rgb.
\end{itemize}

We conducted these experiments to measure the importance of having well converged classification neural network  for each detector.

For the ACTEMIUM dataset, we also trained using the original SSD weights.

All models are evaluated using the common mean Average Precision (mAP) for object detection. The Average Precision for a class is defined as the average of the precision at different recall level (eq \ref{eq:map} and eq \ref{eq:precision}). In the equation, $r$ is the recall level, for the 2007 edition of the PASCAL VOC challenge there were 11 eleven equally spaced recall levels, from 0 to 1. The precision $p()$ is interpolated $p_{interp}(r)$ as the maximum of the precision for all the recall higher than the recall $r$.

\begin{equation}\label{eq:map}
     AP = \frac{1}{11} \sum_{r \in \{0,0.1,...,1\}} p_{interp}(r)
\end{equation}

\begin{equation}\label{eq:precision}
    p_{interp}(r) = \underset{\widetilde{r}: \widetilde{r} \geq r} {\mathrm{max}}p(\widetilde{r})
\end{equation}

\subsection{Results}

\begin{figure}[!ht]
    \centering
    \begin{subfigure}{\includegraphics[width=\columnwidth]{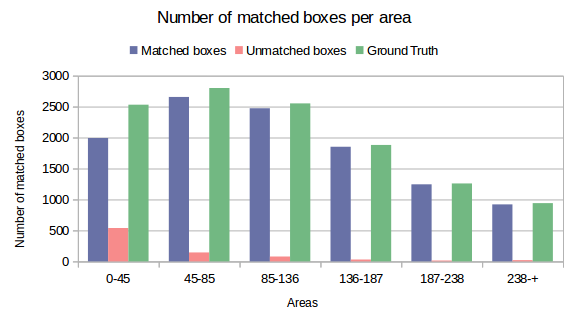}}
       \caption{Results per area for the RGB network retrained from the original VGG16 weights.}
       \label{fig:small_original_rgb} 
    \end{subfigure}

    \begin{subfigure}{\includegraphics[width=\columnwidth]{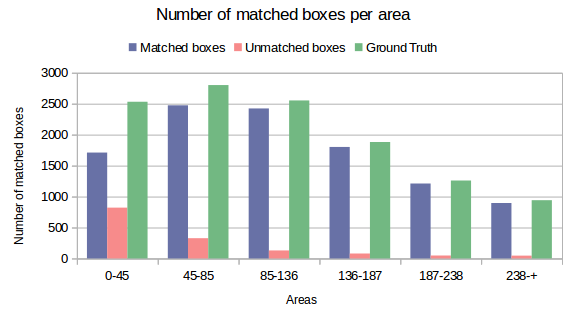}}
        \caption{Results per area for the RGB network retrained from our VGG16 weights.}
        \label{fig:small_retrained_rgb}
    \end{subfigure}

    \begin{subfigure}{\includegraphics[width=\columnwidth]{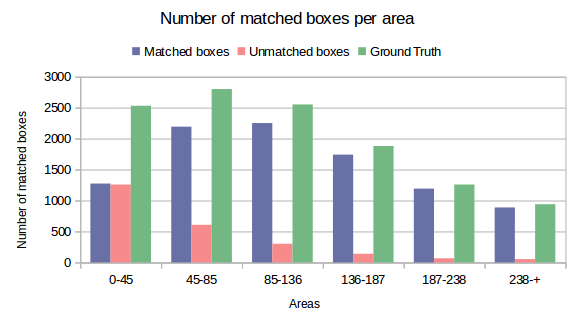}}
        \caption{Results per area for the DCT network retrained from our VGG16 weights.}
        \label{fig:small_dct}
    \end{subfigure}
\end{figure}

We trained on the different combinations of the PASCAL VOC dataset described in the previous section, `07` and `07+12`. Table \ref{table:ssd_results_pascal} reports the obtained detection results on PASCAL VOC data. As expected, the SSD model trained with fully optimized VGG provides the best detection performance. A drop of performance is observed while using the alternative SSD  with partially trained RGB classification module. This fact highlights the influence of initial weights on generalization ability of the detection models. The detection model we design for compressed images provides the lowest mAP. To analyse the rationale of this lower performance, we investigate the observed detection errors with relation to the size of the objects to be identified. This is summarized by the graphs of Figures \ref{fig:small_original_rgb}, \ref{fig:small_retrained_rgb} and \ref{fig:small_dct}. Each graph shows the number of matched object per detection group, i.e $0$ to $45 \times 45$ pixels of area, $45 \times 45$ to $85 \times 85$ pixels of area, etc., for one type of network. The green bar represent the number of object to detect, the blue bar the object matched and the red bar the object unmatched. They clearly show that for small size objects of dimension $45 \times 45$ pixels, the SSD detector on DCT inputs struggles while it matches up the performance of the best RGB-based SSD for large size objects.

Table \ref{table:ssd_results_miisst} gathers observed performances on ACTEMIUM dataset. Compared to Pascal dataset the drop in mAP metrics is slight for the proposed detection model. 
%for We still see a difference between spatial and frequency predictions, but this time the difference much more less significant, about 6.5 points. 
This can be due to the small number of object classes to be identified, hence allowing the SDD network to learn relevant feature maps. 

\begin{figure}[!ht]
    \centering
    \includegraphics[scale=0.38]{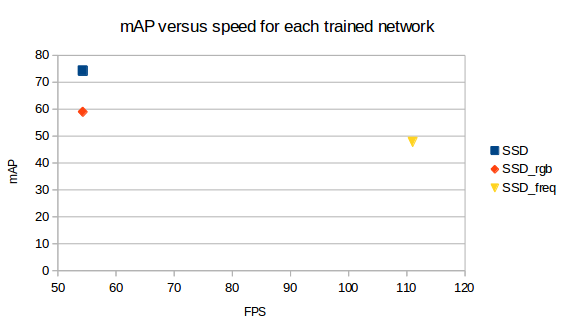}
    \caption{mAP vs speed comparison. We use the speed obtained on the Nvidia GTX. SSD is the original network, SSD\_rgb is the retrained rgb network and SSD\_freq is the DCT trained network.}
    \label{fig:speed_map}
\end{figure}

Finally, we study the impact of our architecture on the network detection speed. First, we roughly estimate the number of FLOPs required for each of the networks, then we carry out speed test on the network on data pre-loaded in memory. We pre-load the data to be sure that the GPU will be used at full capacity. We tested the speed inference by predicting boxes over 619 batches of size 8, reproduced 10 times this experiment and averaged the results for more fairness. All the speed experiments have been performed on a Nvidia GTX 1060 with 6Go of memory. In the end we found the SSD\_freq to be $2.05 \times $ faster than its counter part in spatial domain while keeping good detection performances. We also make experiment to test the loading speed of a DCT JPEG image compare to an RGB one and find a speed improvement of 1.4. Figure \ref{fig:speed_map} compares the speed of our networks with their precision. 

\begin{table}[!ht]
    \centering
    \begin{adjustbox}{max width=\linewidth}
        \begin{tabular}{|c|c|c|c|c|c|c|c|}
        \hline
            Network & From backbone & training dataset & mAP (test 2007) \\
        \hline
            SSD & official VGG & 07+12 & 74.3 \\
            SSD\_rgb & ours & 07+12 & 59.0 \\
            SSD\_freq & ours & 07+12 & 47.8 \\
        \hline
            SSD & official VGG & 07 & 68.0 \\
            SSD\_rgb & ours & 07 & 50.3 \\
            SSD\_freq & ours & 07 & 39.2 \\
        \hline
        \end{tabular}
    \end{adjustbox}
    \caption{Detection results for the training of the SSD on the PASCAL VOC challenge. For the training sets, 07 means Pascal VOC 2007 trainval and 07+12 the union of 2007 and 2012 trainval}
    \label{table:ssd_results_pascal}
\end{table}

\begin{table}[!ht]
    \centering
    \begin{adjustbox}{max width=\linewidth}
        \begin{tabular}{|c|c|c|c|}
        \hline
            Network & From & training dataset & mAP (test set) \\
        \hline
            SSD & official SSD & ACTEMIUM & 82.3 \\
        \hline
            SSD\_rgb & SSD\_rgb (PASCAL) & ACTEMIUM & 77.8 \\
        \hline
            SSD\_freq & SSD\_freq (PASCAL) & ACTEMIUM & 74.6 \\
        \hline 
        \end{tabular}
    \end{adjustbox}
    \caption{Detection results for the training of the SSD on ACTEMIUM dataset.}
    \label{table:ssd_results_miisst}
\end{table}

\begin{table}[!ht]
    \centering
    \begin{adjustbox}{max width=\linewidth}
        \begin{tabular}{|c|c|c|c|c|}
        \hline
            Network & FLOPs(Go) & FLOPs ratio to baseline & FPS & FPS ratio to baseline \\
        \hline
            SSD & 31 & 1 & 54.3 & 1 \\
        \hline
            SSD\_freq & 14 & 2.2 & 111.3 & 2.05 \\
        \hline
        \end{tabular}
    \end{adjustbox}
    \caption{The number of FLOPs required for the two types of network and the corresponding inference speed of the network. All the data was pre-loaded in memory before the tests and the calculation was done on a Nvidia GTX 1060 with 6Go of memory.}
    \label{table:flops}
\end{table}

\section{CONCLUSIONS}

In this article, we have shown that detection is feasible in the JPEG compressed domain, and allow fast detection at the cost of a small accuracy loss, namely regarding small objects. This is a first step toward fast and cheap detection, and many challenges are yet to be addressed. Because of it's design, the JPEG compression algorithm may come with various forms (different sub-sampling ratios, encoding precision, etc.)  which can be handled - although not evaluated - by the proposed approach. Moreover, many techniques exist in the spatial domain to improve speed at the cost of accuracy, it would be interesting to see if all of these techniques are applicable into the frequency domain. We expect the pipeline used for the JPEG images to be applicable to other format using block-wise compression and to yield improvement in the video domain too.


\section*{ACKNOWLEDGMENT}

This research is supported by ANRT and ACTEMIUM Paris Transport. We thank ACTEMIUM Paris Transport for the dataset and the funding. We thank CRIANN for the GPU computation facilities. 

\bibliographystyle{unsrt}
\bibliography{bibliography}

\end{document}